\documentclass{article}
\usepackage{titlesec}
\usepackage{enumitem}
\usepackage[utf8]{inputenc}
\usepackage{natbib}
\usepackage{url} % not crucial - just used below for the URL 
\usepackage{amsmath}
\usepackage{natbib}
\usepackage[colorlinks=true,allcolors=blue]{hyperref}
\usepackage{url}
\usepackage{graphicx}
\usepackage{amsmath, amssymb, gensymb}
\usepackage{bbold}
\usepackage{mathbbol}
\usepackage{algorithm}
\usepackage{algorithmic}
\usepackage{enumitem}
\usepackage{float}
\usepackage{xcolor}
\usepackage{bm} % Math packages
\usepackage{setspace}
\usepackage{multirow}
\usepackage{multicol}
\usepackage{lscape}
\usepackage{rotating}
\usepackage{adjustbox}
\usepackage{booktabs}
\usepackage{ragged2e}

\usepackage{comment} 
\usepackage{subfigure}
\usepackage{tikz}
\usetikzlibrary{positioning, shapes.geometric}
\usepackage{float}

\def\*#1{\mathbf{#1}}
\def\^#1{\mathbf{#1}}
\def\##1{\mathbb{#1}}
\DeclareSymbolFontAlphabet{\amsmathbb}{AMSb}%

\usepackage[a4paper,margin=1in]{geometry}

\title{Probabilistic Classification and Uncertainty Quantification of Sahara Desert Climate Using Feedforward Neural Networks}
\author{Stephen Tivenan, Indranil Sahoo, Yanjun Qian \\ \\
Department of Statistical Sciences and Operations Research, \\
Virginia Commonwealth University
}
\date{}

\begin{document}

\maketitle

\begin{abstract}
Climate classification plays a vital role in agricultural planning, hydrological studies, and climate science. One of the most widely used systems for classifying global climate zones is the K\"oppen-Trewartha (KT) classification. However, the KT classification is fundamentally deterministic, offering discrete labels to spatial locations without accounting for uncertainties in classification.  In this paper, we provide a framework for probabilistic modeling of climatic zones. We implement a feedforward artificial neural network (ANN) for classification, allowing for efficient, uncertainty-aware categorization of climatic regions, thereby offering a more nuanced understanding of transitional climate zones compared to traditional deterministic methods. We apply this method to the Sahara Desert region over the 30-year period of 1960 - 1989, using data at more than 400,000 space-time locations from the first 11 years to train our model. We assess the model's short- and long-term classification capabilities to evaluate its stability and accuracy over time. We also compare the probabilistic classification from our model with the traditional KT classification. In addition, we use fluctuation analysis methods to highlight the temporal evolution of climatic zones across the Sahara region and identify areas undergoing significant flux of probabilities of their climate classes, providing insights into broader trends in desertification.

\end{abstract}

\section{Introduction}
%\noteq{Add a few paragraphs to introduce the definition and importance of climate classification. Talk about desert  climate -- factors affecting desert climate... }

Climate classification can be defined as a categorization method that uses climatic variables to partition an environmental landscape. Climate classification methods have been used for hydrology \citep{knoben2018quantitative, hess-24-4503-2020}, agriculture \citep{Ai_Munandar_2017, hadria2019derivation}, and to study the evolution of climate patterns over time \citep{cui2021observed, he2021assessment, beck2006characterizing}. The purpose of incorporating climate classification is to provide critical insights into environmental conditions that influence a wide range of regional and global phenomena. These include agricultural decisions such as crop selection, water resource management practices, and the prediction of events like droughts, frost, and water runoff. Moreover, climate classification systems facilitate the identification of broader climatic shifts over time, thereby offering valuable information for understanding long-term environmental changes. One of the earliest climate classification systems, the K\"oppen Climate Classification (KCC), was developed using seasonal and monthly precipitation and temperature values to delineate climate regions \citep{koppen1900versuch}. The underlying concept behind the development of climate classifications was the observation that variations in natural vegetation are closely correlated with differences in precipitation and temperature across regions of the Earth \citep{thornthwaite1943problems}. Since its inception, several adaptations of the original KCC have been introduced to refine and improve climate classification globally. The K\"oppen-Geiger Classification (KGC) revises and updates K\"oppen's original system to better represent observed climatic patterns \citep{koppen_geiger_climate_classification}. The K\"oppen-Trewartha Classification (KTC) further modifies the precipitation and temperature thresholds to adjust the criteria for different climate zones \citep{trewartha1943introduction, trewartha1954introduction}. Meanwhile, the K\"oppen-Thornthwaite Classification (KTHC) incorporates additional variables, such as potential evapotranspiration and humidity, to establish more detailed and comprehensive criteria for climate classification \citep{thornthwaite1948approach}. These updated classification systems were designed to better align with human-evaluated climate categories and to fine-tune the representation of diverse climatic regions across the globe \citep{belda2014climate}.

\newpage

Recent studies have increasingly compared traditional climate classification systems to data-driven techniques. A major limitation of the KCC is that it is a rigid, empirical system that lacks the flexibility of dynamic classification approaches \citep{sathiaraj2019predicting}. The KCC also struggles to detect or adapt to changes in regional climates, particularly in areas affected by climate change \citep{triantafyllou1994assessing}. Moreover, because KCC is largely based on static thresholds and expert judgment, it may introduce subjective biases into the classification process. In contrast, newer data-driven approaches offer a more objective framework for categorizing climates by relying on statistical and machine learning techniques. For instance, \citet{netzel2016using} demonstrated that multivariate time series analysis of temperature, precipitation, and temperature range can be employed to assess environmental similarity and delineate climate zones. Similarly, unsupervised clustering algorithms such as K-means, BIRCH, and DBSCAN have been applied to classify climate types based on observed station data across the continental United States \citep{sathiaraj2019predicting}. \citet{zscheischler2012climate} showed that K-means clustering grouped climate data more accurately than traditional KGC and multivariate regression tree methods. Additionally, \citet{lasantha2022data} normalized seasonal, monthly, and annual precipitation data using principal component analysis (PCA) and then applied K-means, ISODATA, and Random Forest clustering methods, comparing the resulting classifications with the KGC. In addition, \citet{sharma2019changes} used probability distributions to analyze global precipitation patterns, and \citet{katzav2021appropriate} discussed the contexts in which probabilistic models are appropriate in climate science. 

However, these approaches are either deterministic or do not explicitly account for uncertainties associated with the predictions. Moreover, the combination of supervised machine learning approaches along with uncertainty quantification has not yet been deeply explored in climate classification. To this end, rather than implementing an unsupervised machine learning approach, we train a feedforward artificial neural network (ANN) for probabilistic climate classification. A feedforward neural network is a specific type of neural network where data flows in one direction, from the input layer, through hidden layers, to the output, without forming cycles. This structure allows feedforward networks to effectively capture complex, nonlinear relationships between inputs and responses, leveraging the interconnected nodes and layers to transform and fit the data with greater flexibility. Although ANNs have been widely applied in environmental research \citep{wikle2023illustration}, the use of feedforward neural networks specifically within climate science has been relatively limited. Notable applications include predicting local microclimates by training on sensor data to forecast temperature and humidity variations based on inputs such as solar radiation and wind speed \citep{zanchi2023harnessing}, modeling intensity of tropical cyclones using data from weather models and observational benchmarks \citep{cloud2019feed}, and forecasting carbon emissions \citep{nie2023novel}. Feedforward neural networks have also been used to estimate soil temperature at varying depths \citep{ozturk2011artificial}, analyze time series data related to climate, particularly in conjunction with transformer architectures \citep{liu2025investigation} and assess the impact of evolving climate on various variables, such as salinity in coastal areas \citep{abiy2022multilayer}. 

This study proposes a novel methodological framework that implements a feedforward ANN for spatial classification to provide probabilistic climate classifications for semi-arid and arid regions. Our ANN model outputs the probability that a given location belongs to a particular climate category in a specific year, thereby directly quantifying the uncertainty associated with climate classification. To our knowledge, previous studies have not explored the use of probability distributions in classifying climate categories.  We apply our methodology to the Sahara and Sahel regions, which offer a rich and challenging testbed due to their diverse climatic and ecological variability. The Sahel region serves as a transitional zone between the Sahara Desert and the African savannah, featuring a complex ecology of shrubland, grassland, wetlands, forest, and desert, with climatic conditions known to fluctuate significantly \citep{wu2022ecological}. In contrast, the Sahara is dominated by more uniform arid conditions, including both arid and hyper-arid zones \citep{brito2021drivers}. For our analysis, we classify the Sahara and Sahel regions into three categories, namely, arid (desert), semi-arid (steppe), and non-arid (non-desert), for the period spanning 1960 to 1989. Each pixel’s classification is based on a continuous metric ($P/R$), derived from KTC thresholds and a set of spatial and temporal basis functions. After classification, we compare each pixel's new categorization with the original KTC classification and assess the associated uncertainty for each category, thereby providing insights into regions with strong versus weak probabilities for a particular climate categorization.  

Thus, our study introduces the feedforward artificial neural network as an efficient and flexible tool for climate classification. Specifically, we demonstrate its potential to (1) predict arid and semi-arid classifications with greater adaptability and (2) provide explicit probability and uncertainty estimates alongside established classifications. By incorporating uncertainty into the classification process, we highlight that the boundaries between climate categories are often more nuanced and dynamic than traditional deterministic approaches suggest. We therefore offer a deeper probabilistic perspective into the complex and evolving relationships between climatic regions, providing a valuable complement to existing classification frameworks. 

The rest of the paper is organized as follows. Section \ref{data} gives a description of the dataset used in our study. Section \ref{methods} discusses the background of feedforward neural networks and the proposed model. Section \ref{results} shows the results obtained from the proposed model and comparisons with the KTC. Finally, a discussion of the results is presented in Section \ref{dis}, and some implications and conclusions are presented in Section \ref{con}.

\section{Data Description}  \label{data}
%\subsection*{Data Description}
% Add maps of precipitation and temperature

% What is the variable that are being selected, mentioned how many pixels that we have and then mention how many years we have and in total we have to show how many in total data we have 

% Write about the repository, finding the to

During the 1970s and 1980s, the Sahara and sub-Sahara regions experienced a series of devastating droughts \citep{NicholsonSharonE.2018RotA}. These droughts led to severe consequences, including mass migration, overgrazing, and significant socio-economic challenges across the region \citep{hess-18-3635-2014}. Agricultural and irrigation systems were heavily strained, while the biosphere suffered widespread degradation \citep{glantz1987drought}. In response to these crises, the United Nations initiated financial and humanitarian support through Desertification Conventions held in 1977 and 1992 \citep{kassas1995desertification}. Additionally, the Sahara and Sahel Observatory (OSS) established the Long-Term Ecological Monitoring Network (R\'eseau d'observatoires de surveillance \'ecologique \`a long terme, ROSELT/OSS) to monitor climatic variables and provide critical environmental information to regional governments to combat land degradation \citep{VogtJ.V.2011Maao}. Despite these efforts, the Sahara-Sahel zone remains among the most vulnerable areas to desertification. Projections indicate that the region will face increasing risk due to a combination of economic instability, agricultural dependency, poverty, and the intensification of arid conditions \citep{HuangJianping2020Gdvt}. Given the critical importance of the region and the profound climatic impacts during the 1970s and 1980s, we selected the spatio-temporal domain from 1960 to 1989 in the Sahara-Sahel area as the focus of our study.

The dataset used in this study is obtained from the GLDAS Noah Land Surface Model Version 2.0, a global land data assimilation system (GLDAS) that provides data on various climatic and temporal variables from 1948 to 2014 \citep{beaudoing_rodell_2020}. GLDAS Version 2.0 offers data at both three-hourly and monthly time scales, where the monthly values represent averages over all three-hourly records within each month. From this dataset, we selected two monthly variables: total precipitation rate (measured in $\textrm{kg}~\textrm{m}^{-2}~\textrm{s}^{-1}$) and instantaneous air temperature (measured in Kelvin, K). For the purposes of our analysis, these variables were converted to Celsius ($\degree$C) for temperature and centimeters (cm) for precipitation.

To obtain total monthly precipitation in centimeters, the precipitation rate was first converted to eliminate the time dimension. This involved multiplying by the total number of seconds in a month, accounting for the number of days ($n$), and then applying appropriate unit conversions, as shown below:
\begin{equation*} \frac{1~\textrm{kg}}{1~\textrm{m}^2~\textrm{s}} \times \frac{10800 ~\textrm{s}}{3 ~\textrm{hours}} \times 8 \left(\frac{3 ~\textrm{hours}}{1~\textrm{day}}\right) \times n~\textrm{days} = \frac{\textrm{kg}}{\textrm{m}^2}. \end{equation*}
Then, the second step is to use the density of the water to convert mass per unit area to centimeters:
\begin{equation*} \frac{1~\textrm{kg}}{1~\textrm{m}^2} \times \frac{1~\textrm{m}^3}{1000~\textrm{kg}} \times \frac{100~\textrm{cm}}{1~\textrm{m}} = 0.1~\textrm{cm}. \end{equation*}
For temperature, no temporal adjustment was necessary since the instantaneous air temperature variable already represented the average across all three-hourly intervals for a given month. Both precipitation and temperature variables were generated by GLDAS through the Princeton meteorological forcing dataset \citep{sheffield2006development}, which assimilates observational data to provide consistent inputs across the time period.

\begin{figure}[ht] 
    \centering
    \subfigure[Precipitation 1975]{\includegraphics[width=0.45\textwidth]{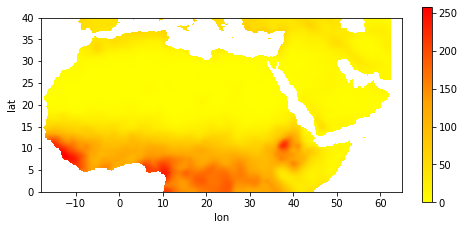}
    }
    \centering
    \subfigure[Temperature 1975]{
        \includegraphics[width=0.45\textwidth]{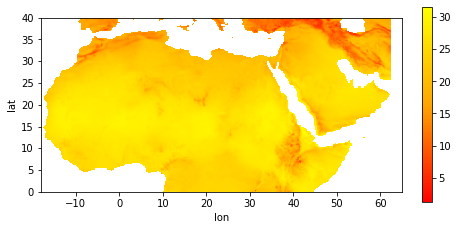}
    }
    \subfigure[Precipitation 1980]{
        \includegraphics[width=0.45\textwidth]{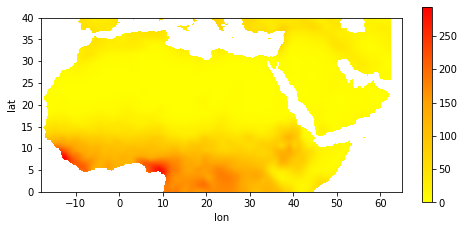}
    }
        \subfigure[Temperature 1980]{\includegraphics[width=0.45\textwidth]{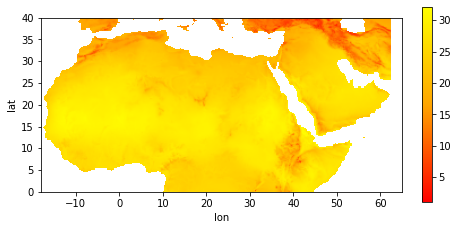}
    }
    \centering
    \subfigure[Precipitation 1985]{
        \includegraphics[width=0.45\textwidth]{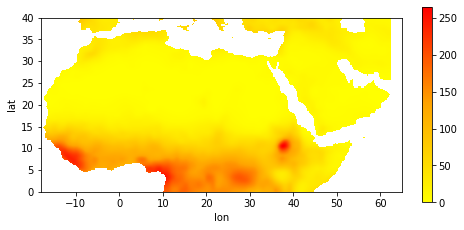}
    }
    \subfigure[Temperature 1985]{
        \includegraphics[width=0.45\textwidth]{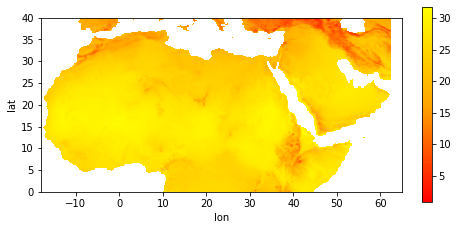}
    }
    \caption{Spatial distribution of total precipitation and average temperature (in centimeters and Celsius) over the Sahara and Sahel regions for the years 1975, 1980, and 1985. The maps illustrate the variability of precipitation across the study area over the selected years, highlighting regional differences in climatic conditions.}
    \label{fig:Precipitation}
\end{figure}

The spatial resolution of GLDAS 2.0 varies between 0.25$\degree$ and 1$\degree$. In our analysis, we used the 0.25$\degree$ resolution data for both variables, covering the entire Sahara and Sahel regions from 0$\degree$N to 40$\degree$N latitudes and from -20$\degree$W to 60$\degree$E longitude, on a monthly basis from 1960 to 1989. This selection resulted in approximately 36,745 data points for each time point. Figure \ref{fig:Precipitation} shows the spatial distribution of total precipitation (in centimeters) across the Sahara and Sahel regions for three candidate years, 1975, 1980, and 1985, within our temporal domain. The maps illustrate the variability of precipitation across the study domain over time, highlighting regional differences in climatic conditions.

% ADD SPATIAL MAPS OF TEMPERATURE IN Celcius.

While the K\"oppen-Trewartha Classification (KTC) system has been a reliable framework to assess and project global climate simulations \citep{remedio2019evaluation, BeldaMichal2015EoCp, belda2014climate, BeldaMichal2016Gwci}, a central theme among these studies has been the link between observed and projected climate shifts and broader climate trends. In contrast to these studies, we do not use the KTC to suggest climatic transitions, but rather to serve as a ground truth for training and evaluating our model, specifically trained to distinguish between two dry climate types: arid/desert and semi-arid/steppe. 

Following the KTC framework for dry climates, we classify each spatial location into three categories: arid (desert), semi-arid (steppe), or non-arid. The classification is based on annual precipitation totals and Patton’s precipitation threshold \citep{patton1962note}, denoted as $R$, for a given year. The threshold $R$ is calculated as:
\begin{equation*} R = 2.3T - 0.64P_W + 41, \end{equation*}
where $T$ is the mean annual temperature in degrees Celsius ($\degree$C), and $P_W$ is the percentage of total annual precipitation that occurs during the winter months (October - March, in the Northern Hemisphere). Based on this threshold, the classification rule for dry climates is as follows:
\begin{equation*} \text{Classification} = \begin{cases} \text{Arid}, & \text{if } P < R/2, \\ \text{Semi-Arid}, & \text{if } R/2 \leq P < R, \\ \text{Non-Arid}, & \text{if } P \geq R, \end{cases} \end{equation*}
where $P$ denotes the mean annual precipitation. For each year from 1960 to 1989, we use aggregated precipitation and temperature data to compute the KTC-based classification at each spatial location. This process results in each pixel being labeled as arid, semi-arid, or non-arid for each year within our study period. The variable $P/R$ serves as a key variable in our modeling framework, providing an interpretable and climate-informed basis for supervised learning. By using $P/R$ as a continuous variable rather than relying solely on rigid discrete classifications, we capture finer local variations across space and time. This approach allows us to observe fluctuations over time and identify deviations from the standard KTC based on the established criteria. 

\begin{figure}[]
    \centering
    \includegraphics[width=0.9\linewidth]{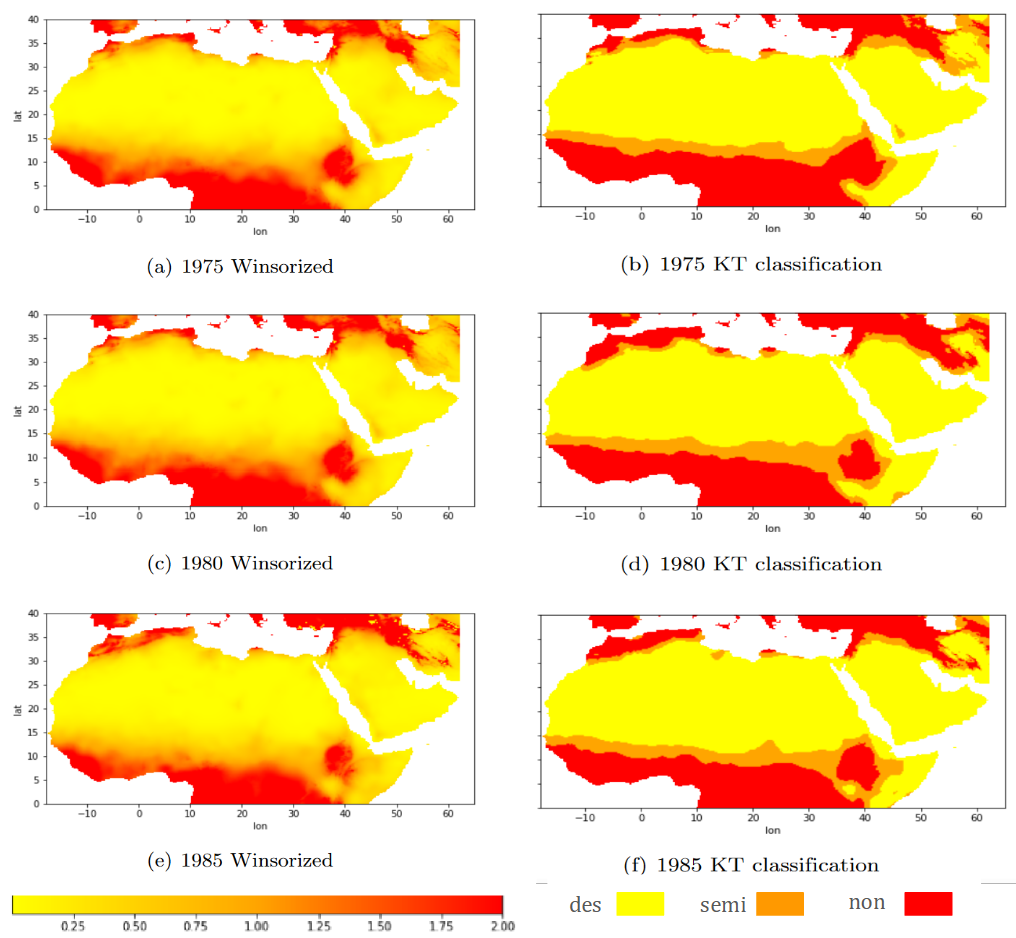}
    \caption{Spatial maps of winsorized $P/R$ values (left column) and corresponding KTC dry climate classifications (arid, semi-arid, and non-arid; right column) for the candidate years 1975, 1980, and 1985. The $P/R$ values are winsorized between 0.001 and 2 to enhance visualization clarity.}
    \label{fig:PoverR}
\end{figure}

Figure \ref{fig:PoverR} displays the spatial distribution of winsorized  $P/R$ values across the Sahara and Sahel regions for the years 1975, 1980, and 1985, along with the corresponding KTC systems dry climate classifications. The raw $P/R$ values exhibited a wide range, including extremely large and even negative values, which obscured meaningful spatial patterns if plotted directly. Hence, we winsorized the $P/R$ values to lie between 0.001 and 2 for clear visualizations. In the maps, the relationship between the spatial variation of $P/R$ and the classified arid, semi-arid, and non-arid regions can be easily observed.

%\subsection{Data Classification}

\section{Feedforward Neural Network Modeling Framework} \label{methods}

\subsection{Model Architecture}
Let $(\bm{s}, t)$ denote spatio-temporal locations with $(\bm{s}, t) \in \mathcal{D} = \lbrace (\bm{s}, t): \bm{s} = \bm{s}_1, \ldots, \bm{s}_N \text{ and } ~t = 1, \ldots, T\rbrace$ and define
\begin{equation*} Y(\bm{s}, t) = \begin{cases} 1, & \text{if location } (\bm{s}, t) \text{ is arid}, \\ 2, & \text{if location } (\bm{s}, t) \text{ is semi-arid}, \\ 3, & \text{if location } (\bm{s}, t) \text{ is non-arid} \end{cases} \end{equation*} with $\mbox{P}\left[ Y(\bm{s}, t) = c \right] = p_c(\bm{s}, t), c = 1, 2, 3$. In this section, we introduce a spatio-temporal feedforward neural network model with $L$ layers to estimate class probabilities $p_c(\bm{s},t)$ for categorical responses that vary over space and time.

Let $\bm{X}(\bm{s}, t)$ denote the $p \times 1$ set of covariates observed at the spatio-temporal location $(\bm{s}, t)$. To accurately capture the spatial structure of the process, we incorporate spatio-temporal basis functions as inputs to the network, following a strategy similar to the work of \cite{NagPratik2023SDfi}, as opposed to directly feeding the coordinates of the observations as input to the network. We use the multi-resolution compactly supported Wendland radial basis functions \citep{nychka2015multiresolution} as our spatial basis functions. The Wendland basis functions are defined as 
$$
B_1(d) = \frac{(1-d)^{6}}{3} (35d^{2} + 18d + 3), ~0 \leq d \leq 1  
$$
and the spatial basis functions are then defined as 
$$
\phi_i(\bm{s}) = B_1\left( \frac{||\bm{s} - \bm{u}_i||}{\theta}\right),
$$
where $\lbrace \bm{u}_1, \ldots, \bm{u}_k\rbrace$ denotes a set of $k$ knot locations and $\theta$ is the bandwidth parameter. We choose $\lbrace \bm{u}_1, \ldots, \bm{u}_k\rbrace$ to be a square grid of varying sizes and $\theta$ to be 2.5 times the maximum distance among all knot locations. The temporal dependence is extracted using Gaussian radial basis functions given by 
$$
\psi_j(t) = \exp\left(\frac{(t - v_j)^{2}}{2\kappa}\right)
$$
with equidistant temporal knots  $\lbrace v_1, \ldots, v_r\rbrace$, and $\kappa = |v_1 - v_2|$. Let 
$$\bm{B}(\bm{s}, t) = \lbrace \phi_1(\bm{s}), \ldots, \phi_k(\bm{s}), \psi_1(t), \ldots, \psi_r(t)\rbrace^T$$
denote the stacked vector of length $(k + r)$ of spatial and temporal basis functions at $(\bm{s}, t)$. The stacked spatial and temporal basis functions are included as inputs to the feedforward neural network, allowing the model to flexibly capture the spatio-temporal structure of the data. This significantly reduces the computational burden typically associated with modeling spatial dependence through complex parametric covariance structures. This approach does not imply a separable spatio-temporal model; although the bases are stacked, the neural network learns shared weights across nodes, enabling it to model interactions between spatial and temporal components and thereby capture non-separable spatio-temporal dependencies. 

Let $\bm{X}_{vec}(\bm{s}, t) = \lbrace \bm{X}(\bm{s}, t)^T, \bm{B}(\bm{s}, t)^T\rbrace^T$ denote the embedded input vector of length $(p + k + r)$ at location $(\bm{s}, t)$. Given the input variables $\bm{X}_{vec}(\bm{s}, t)$, our feedforward neural network with $L$ layers and $M_l$ nodes in layer $l$ is given as
\begin{align*}
    h_i^{l}(\bm{s}, t) = \sigma \left( \sum_{j = 1}^{M_{l- 1}} w^l_{ij} h_{j}^{l - 1}(\bm{s}, t) + b_{i}^l\right),
\end{align*}
for $l = 2, \ldots, L$ with $h_i^{1}(\bm{s}, t) = \bm{X}_{vec; ~i}(\bm{s}, t)$, the $i^{th}$ input variable at $(\bm{s}, t)$ and $M_L = 3$. Here, $w^l_{ij}$ is the weight from the $j^{th}$ node in the $(l - 1)^{th}$ layer to the $i^{th}$ neuron in the $l^{th}$ layer.  The term $b_{i}^l$ is the bias from the $i^{th}$ node in the $l^{th}$ layer and $\sigma(\cdot)$ denotes a possibly nonlinear activation function, which in our case is the rectified linear unit (ReLU) \citep{schmidt2020nonparametric}. In our application, the feedforward neural network architecture has been built using the \texttt{keras} package \citep{chollet2015keras} in Python. To reduce overfitting the model to the training data, we set a dropout rate for our hidden layers, which randomly sets a specified percentage of the layer's weights to zero during training. 

\begin{figure}[]
    \centering
    \includegraphics[width=.75\linewidth]{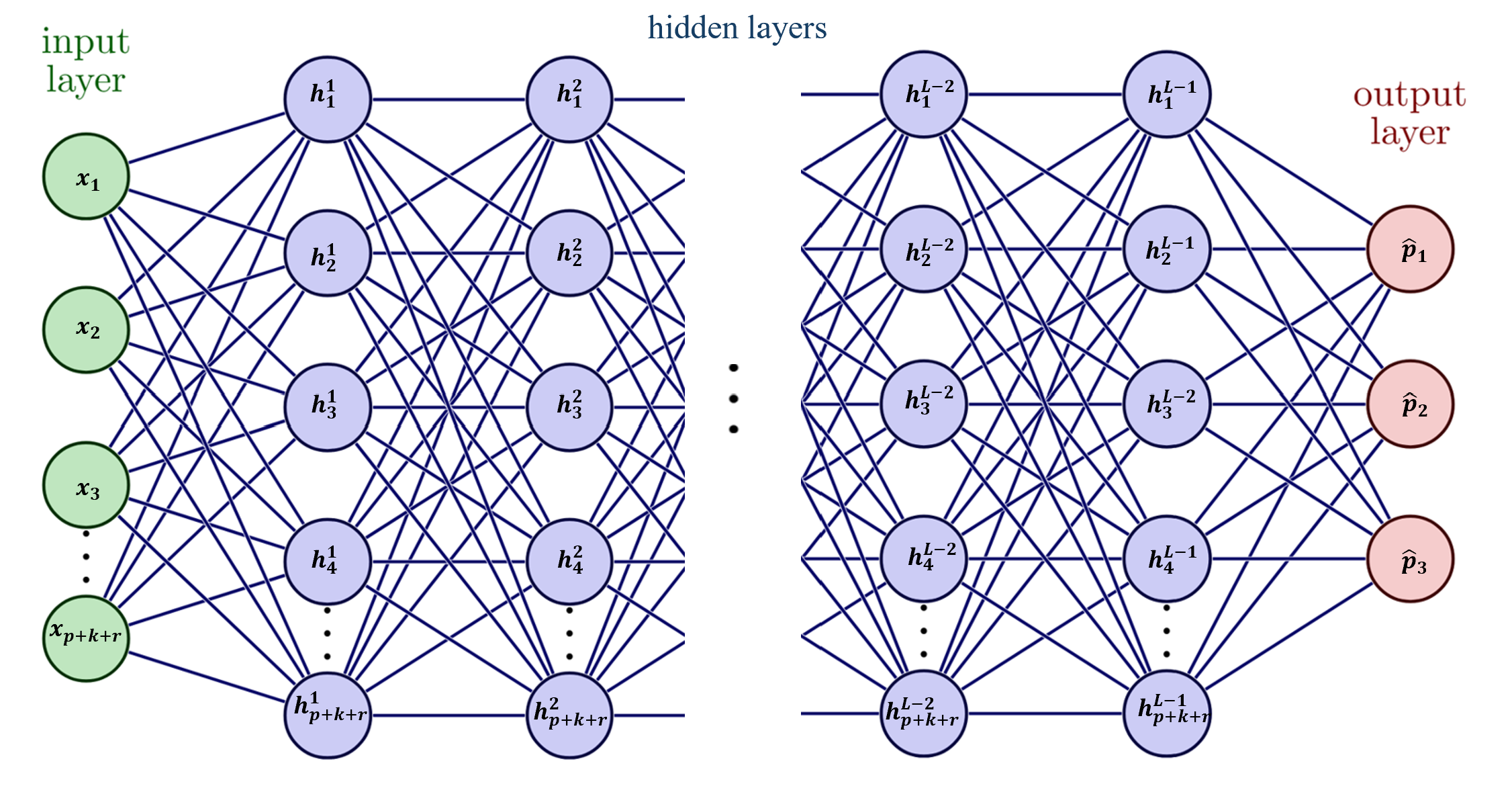}
    \caption{Architecture of the feedforward neural network for spatio-temporal classification at location $(\bm{s}, t)$. This example network has $L - 1$ hidden layers, each with $p + k + r$ nodes.}
    \label{fig:FNN}
\end{figure}

The activation function used in the final layer of the neural network is the softmax function, which outputs the probabilities associated with each of the three climate categories at location $(\bm{s}, t)$ as follows: 
$$
\widehat{p}_c(\bm{s}, t) = \mbox{softmax}\lbrace h^L_c(\bm{s}, t); ~c = 1, 2, 3 \rbrace = \frac{e^{h_c^L(\bm{s}, t)}}{\sum_{c = 1}^3 e^{h_c^L(\bm{s}, t)}},
$$
for $c = 1, 2, 3$. A depiction of this architecture is given in Figure \ref{fig:FNN}.

The weights and biases in the neural network are estimated from the training data by minimizing the sparse categorical cross-entropy (SCCE) loss function, which evaluates the model fit by comparing the predicted probabilities with the known categorical labels. Specifically, in the training batch, the SCCE computes the sum of the product of the true category indicator $Y(\bm{s}, t)$ and the predicted probability $\widehat{p}_c(\bm{s}, t), ~c = 1, 2, 3$ for the three classification categories across all locations:
$$
\text{SCCE} = - \sum_{t = 1}^{T_r}\sum_{i = 1}^N \sum_{c = 1}^3 \mathbb{1}\lbrace Y(\bm{s}_i, t)=c\rbrace \widehat{p}_c(\bm{s}_i, t), 
$$
where $T_r$ represents the number of time points in the training set, and $\mathbb{1}\lbrace \cdot \rbrace$ denotes an indicator function. The network then employs back-propagation to update the weights and biases by minimizing the SCCE loss. This iterative process adjusts the model parameters to achieve the lowest possible loss and improves the sparse categorical cross-entropy accuracy. Here, we implement the `adam' optimizer \citep{kingma2014adam}, an adaptive stochastic gradient descent method, to train our neural network.

\subsection{Model Implementation}

The proposed feedforward neural network model consists of an input layer, two hidden layers, and an output layer. The input features include the $P/R$ metric, spatial basis functions constructed using Wendland kernels, and temporal basis functions defined using Gaussian kernels, as described in earlier sections. The total number of input features depends on the chosen configuration of spatial and temporal basis functions. For example, a $5 \times 5$ grid of spatial basis functions combined with 5 temporal basis functions and the $P/R$ metric yields 31 input features.

We have evaluated multiple configurations of spatial ($5 \times 5$, $7 \times 7$, and $12 \times 12$) and temporal (5, 7, and 10) basis functions and found that the $5 \times 5$ spatial and 5 temporal basis function combination resulted in the best predictive performance. Knot locations for both spatial and temporal bases were selected to be equally spaced across the longitude–latitude grid and the relevant time interval, respectively.

Each hidden layer employs the ReLU activation function and includes a dropout layer with a dropout rate of 50\% to prevent overfitting. The final output layer uses a softmax activation function to estimate the probability of each climate category (arid, semi-arid, and non-arid). The model was trained using data at over $400,000$ spatial locations from 1960 to 1970 and validated on a separate test set from 1971 to 1989. The `adam' optimizer's hyperparameters were set to their standard default values: learning rate $\nu = 0.001$, exponential decay rates $\beta_1 = 0.9$ and $\beta_2 = 0.999$, and a small constant $\epsilon = 10^{-8}$ for numerical stability.

\subsection{Model Performance}
To evaluate the performance of our fitted model on the test set, we use the trained feedforward neural network to estimate the probabilities of each location in each year falling into one of the three climate categories: arid, semi-arid, and non-arid. Each location is then assigned to the category with the highest predicted probability, denoted by $\widehat{Y}(\bm{s}, t)$. Taking the KTC as the ground truth, we compute standard classification metrics such as Precision, Recall, and F1 Score, based on the counts of true positives (TP), false positives (FP), false negatives (FN) and true negatives (TN).

In the context of a three-class classification problem, these metrics are computed for each class using a one-vs-rest approach. For a given class, true positives (TP) are instances correctly predicted as that class; false positives (FP) are instances incorrectly predicted as that class; false negatives (FN) are instances of that class predicted as something else; and true negatives (TN) are all other correctly predicted instances not belonging to that class. That is, for class $c$ and year $t$,
\begin{align*}
    TP_{c, t} &= \sum_{i = 1}^N \mathbb{1} \Bigl\{ Y(\bm{s}_i, t) = \widehat{Y}(\bm{s}_i, t) = c \Bigr\} \\
    FP_{c, t} &= \sum_{i = 1}^N \mathbb{1} \Bigl\{ Y(\bm{s}_i, t) \neq c \text{ and } \widehat{Y}(\bm{s}_i, t) = c \Bigr\} \\
    FN_{c, t} &= \sum_{i = 1}^N \mathbb{1} \Bigl\{ Y(\bm{s}_i, t) = c \text{ and } \widehat{Y}(\bm{s}_i, t) \neq c \Bigr\} \\
    TN_{c, t} &= \sum_{i = 1}^N \mathbb{1} \Bigl\{ Y(\bm{s}_i, t) = \widehat{Y}(\bm{s}_i, t) \neq c \Bigr\},
\end{align*}
where $\mathbb{1}\lbrace \cdot \rbrace$ denotes an indicator function. These values are then used to compute the evaluation metrics separately for each class, $c$ and each prediction year, $t$ as follows:
\begin{align*}
%\text{Accuracy}_{c, t} &= \frac{TP_{c, t} }{TP_{c, t} + TN_{c, t} + FP_{c, t} + FN_{c, t}} \\
\text{Precision}_{c, t} &= \frac{TP_{c, t}}{TP_{c, t} + FP_{c, t}} \\
\text{Recall}_{c, t} &= \frac{TP_{c, t}}{TP_{c, t} + FN_{c, t}} \\
\text{F1score}_{c, t} &= 2 \times \frac{\text{Precision}_{c, t} \times \text{Recall}_{c, t}}{\text{Precision}_{c, t} + \text{Recall}_{c, t}}.
\end{align*}

%\noteq{Qian: I am confused by the definition of $\text{Accuracy}_{c, t}$. The numerator is the number of locations with $Y(\bm{s}_i, t) = \widehat{Y}(\bm{s}_i, t)$ regardless of $c$, and the denominator is the number of all locations. In this case, all $c=1,2,3$ should have the same accuracy. Could you confirm?}

\section{Results}\label{results}
\subsection{Classification Metrics}

We applied the feedforward neural network using climate data from 1960 to 1970 as the training set and data from 1971 to 1989 as the test set. Various configurations of spatial and temporal basis functions were tested as inputs to evaluate the model’s ability to classify each pixel into one of three climate categories: arid, semi-arid, or non-arid. Among the tested models, the configuration with $5\times 5$ spatial knots and $5$ temporal knots was selected as the final model due to its strong performance across all categories (each exceeding $80\%$ accuracy scores) and its relatively simple structure. Simpler models are also less prone to overfitting, and we observed that increasing the number of input parameters often led to reduced accuracy. The model took 8 minutes and 12 seconds to train using data at over 400,000 spatio-temporal locations on a personal laptop with specifications of 12th Gen Intel(R) Core(TM) i7-1260P, 2100 Mhz, 12 Core(s) and 16 Logical Processor(s).

\begin{figure}
    \centering
    \includegraphics[width = \textwidth]{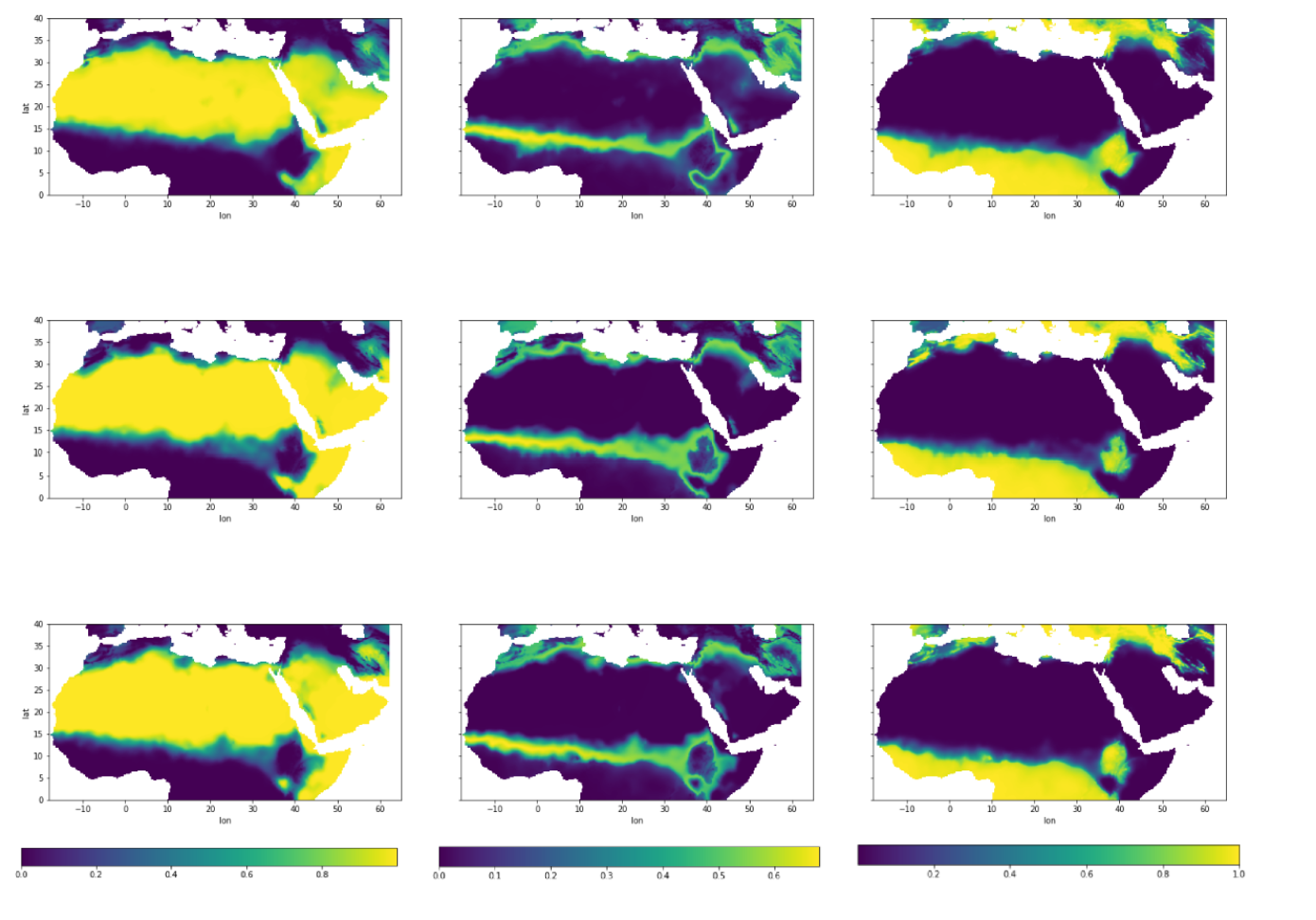}
    \caption{Predicted probability maps for arid, semi-arid, and non-arid regions over the Sahara and Sahel, generated using the fitted feedforward neural network model. Each row represents one of the candidate years (1975, 1980, and 1985), while each column corresponds to one of the three climate categories (arid, semi-arid, and non-arid).}
    \label{fig:Probability Map}
\end{figure}

Using the fitted model, we generated pixel-wise category probabilities for each year from 1971 to 1989. Figure \ref{fig:Probability Map} shows the predicted probability maps for arid, semi-arid, and non-arid regions across the Sahara and Sahel for the years 1975, 1980, and 1985, derived from the fitted feedforward neural network model. Each row corresponds to one of the candidate years, while the columns represent the probability of a pixel belonging to the arid, semi-arid, and non-arid category, respectively. The plots for all three years display a consistent spatial pattern. The central Sahara, including parts of Algeria, Libya, Egypt, Mauritania, Mali, and northern Chad, shows high probabilities for the arid class, aligning with its well-established hyper-arid climate. The semi-arid band, stretching across countries like southern Mauritania, northern Senegal, central Mali, Niger, northern Nigeria and parts of Sudan, reflects the transitional Sahel zone that has historically experienced high inter-annual variability in rainfall and vegetation cover. This pattern is supported by prior climatological and ecological studies, which identify the Sahel as a climatically sensitive region straddling the desert and savanna zones \citep{hulme2001climatic, nicholson2013west}. The non-arid category appears prominently in southern Sudan, Ethiopia, parts of Nigeria and the Guinea Coast, consistent with regions receiving relatively higher and more stable rainfall.

\begin{figure}
    \centering
    \subfigure{
        \includegraphics[width=0.45\textwidth]{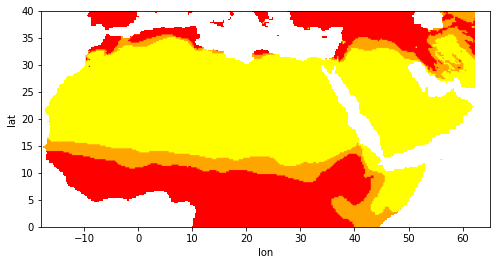}
    }\subfigure{
        \includegraphics[width=0.45\textwidth]{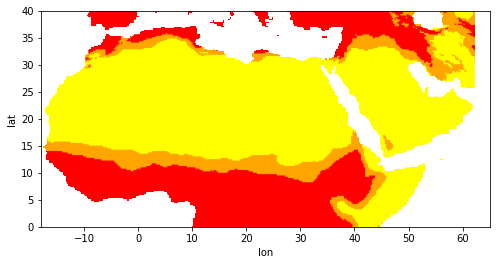}
    }\\
    \subfigure{
        \includegraphics[width=0.45\textwidth]{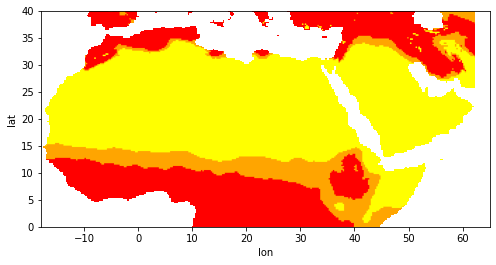}
    }\subfigure{
        \includegraphics[width=0.45\textwidth]{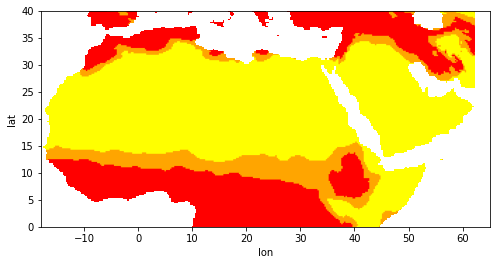}
    }\\
    \subfigure{
        \includegraphics[width=0.45\textwidth]{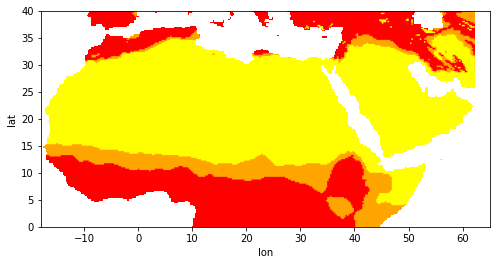}
    }\subfigure{
        \includegraphics[width=0.45\textwidth]{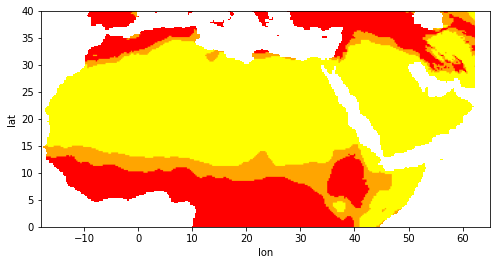}
    }
     \caption{Comparison of climate classifications obtained from the proposed neural network-based method (left column) and the KT classification (right column) for the years 1975, 1980, and 1985. Each row corresponds to one of the candidate years, highlighting the spatial distribution of arid, semi-arid, and non-arid zones across the Sahara and Sahel regions.}
    \label{fig:Predicted_Maps}
\end{figure}

Each pixel was classified into the category with the highest predicted probability. Figure~\ref{fig:Predicted_Maps} presents the predicted climate classifications for the years 1975, 1980, and 1985 alongside the corresponding KT classification maps. To evaluate the performance of the fitted model in predicting arid, semi-arid, and non-arid regions over the Sahara and Sahel, we calculate the precision, recall, and F1-score metrics, using the KT classification as the reference. Table \ref{tab:classification_metrics} presents these classification metrics for the years 1975, 1980, and 1985.  Overall, the model shows excellent performance in classifying the arid and non-arid regions across all years, with precision, recall, and F1-scores consistently above $95\%$ in these categories. For instance, the arid region exhibits F1-scores of $98\%$, $98\%$, and $97\%$ for 1975, 1980, and 1985, respectively, indicating near-perfect agreement with the KT classification.

In contrast, the semi-arid category shows relatively lower performance across all metrics and years. The F1-scores for semi-arid regions are 81\%, 80\%, and 78\% for 1975, 1980, and 1985 respectively, suggesting that the model has more difficulty distinguishing semi-arid zones from the neighboring arid and non-arid zones. This may be attributed to the transitional and ecologically variable nature of the semi-arid belt, particularly across the Sahel, where climate vary rapidly over space and time. Such challenges in classification have been acknowledged in earlier studies \citep{hulme2001climatic, nicholson2013west}, which document the semi-arid Sahel as a climatically unstable region prone to oscillations between aridification and greening depending on rainfall variability. A plot of the precision, recall, and F1-score metrics for all time points across the arid, semi-arid, and non-arid categories is provided in Appendix \ref{app1}.

\begin{table}
\centering
\caption{Classification Metrics for the Years 1975, 1980, and 1985}
\label{tab:classification_metrics}
\begin{tabular}{|c|c|c|c|c|}
\hline
\textbf{Year} & \textbf{Metric} & \textbf{Non-arid} & \textbf{Semi-arid} & \textbf{Arid} \\
\hline
\multirow{3}{*}{1975} 
& Precision & 95\% & 81\% & 98\% \\
& Recall    & 97\% & 82\% & 97\% \\
& F1-score  & 96\% & 81\% & 98\% \\
\hline
\multirow{3}{*}{1980} 
& Precision & 99\% & 80\% & 96\% \\
& Recall    & 92\% & 80\% & 99\% \\
& F1-score  & 96\% & 80\% & 98\% \\
\hline
\multirow{3}{*}{1985} 
& Precision & 99\% & 80\% & 95\% \\
& Recall    & 93\% & 76\% & 99\% \\
& F1-score  & 96\% & 78\% & 97\% \\
\hline
\end{tabular} 
\end{table}

\subsection{Fluctuation Analysis}
To quantify the temporal fluctuation in the climate zones classifications, we implemented a coefficient of variation (CV) analysis, a well-established approach for capturing variability and instability over time \citep{kesteven1946coefficient, wu2020spatial, wu2022ecological, wang2023insights}. In our study, we used the predicted probabilities to first calculate the pixel-wise average probabilities of each climate class across the test set years of 1971 to 1989. For each pixel, the aridity class with the highest average probability over time was identified, and this dominant class served as the baseline for fluctuation analysis at that location.

\begin{figure}[!htpb]
    \centering
        \includegraphics[width = 0.9\textwidth]{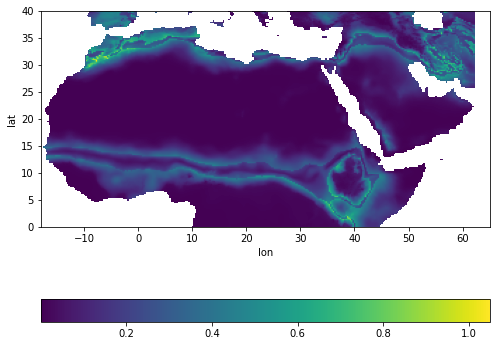}
    \caption{Spatial distribution of the coefficient of variation (CV) of the dominant predicted aridity classification probability from 1971 to 1989 across North and East Africa and the Middle East. Higher CV values indicate greater temporal fluctuation in the dominant aridity classification, especially in transitional zones such as Northwestern Africa, the Sahel, the Horn of Africa, and areas east of Saudi Arabia.}
    \label{fig:Coefficient of Variability}
\end{figure}

We then computed the CV for each pixel, defined as the ratio of the standard deviation to the mean of the class probability for the dominant aridity category over 1971 to 1989. This CV quantifies the relative variability in classification certainty over time. A higher CV indicates greater volatility or fluctuation in the classification for that location, while a lower CV implies temporal consistency in the predicted climate type \citep{wu2020spatial}. Figure \ref{fig:Coefficient of Variability} shows the spatio-temporal fluctuations of the predicted classification probabilities, highlighting evident spatial heterogeneity. We observe that regions with the highest variability correspond to transitional zones, where the boundaries between aridity classifications are more fluid over time. In particular, the higher CV values were concentrated along the Sahel belt, Northwestern Africa, particularly along the Atlas mountain range and the northern Sahel, the Horn of Africa, particularly Ethiopia and Somalia, and the region east and northeast of Saudi Arabia. 

To reflect the fluctuations in climatic zones for intuitively, the CV values were divided into five fluctuation levels, namely, very low fluctuation, low fluctuation, moderate fluctuation, high fluctuation, and very high fluctuation \citep{wang2018spatio}, as shown in Table \ref{tab: Fluctuation degree}. The area with less fluctuation accounted for the largest proportion, covering 65.65\% of the study region. Areas with low and moderate fluctuation were also substantial, representing 17.37\% and 10.86\% of the region, respectively. Areas with high fluctuation and very high fluctuation comprised only 4.37\% and 1.75\% of the region, respectively. The highest CV values were mainly concentrated in ecological transition zones such as the Horn of Africa, the Sahel, northwest Africa, and parts of eastern Saudi Arabia, indicating greater fluctuation in aridity classification over time. Meanwhile, the areas with low or less fluctuation were predominantly located in climatically consistent zones such as the Sahara Desert and the interior of the Arabian Peninsula, suggesting a stable temporal pattern in aridity classification.

\begin{table}[!htpb]
    \centering
    \caption{Fluctuation levels of the coefficient of variation (CV) values in the study region}
    \label{tab: Fluctuation degree}
    \begin{tabular}{|c|c|c|}
    \hline
      \textbf{CV value} & \textbf{Fluctuation Level} & \textbf{Area proportion (\%)} \\ \hline 
         CV $ \leq  0.1$ & Very low fluctuation & 65.65 \\
         $0.1 < $ CV $\leq 0.2$ & Low fluctuation & 17.37 \\
         $0.2 < $ CV $\leq 0.3$ & Moderate fluctuation & 10.86 \\
         $0.3 < $ CV $\leq 0.4$ & High fluctuation & 4.37 \\
         CV $> 0.4$ & Very high fluctuation & 1.75 \\ \hline
    \end{tabular}
\end{table}

\begin{figure}
    \centering
    \subfigure[Ethiopia]{
        \includegraphics[width=0.45\textwidth,  height=.30\textheight]{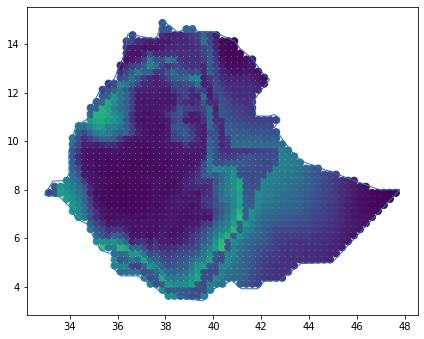}
    }
    \subfigure[Morocco]{
        \includegraphics[width=0.45\textwidth,  height=.30\textheight]{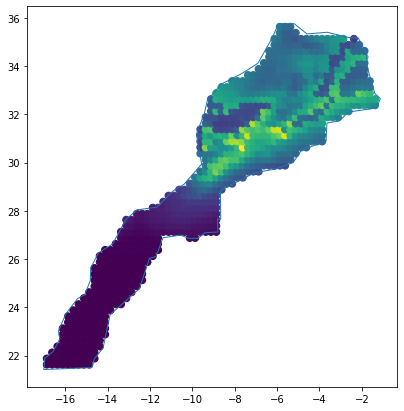}
    }
    \subfigure[South Sudan]{
        \includegraphics[width=0.45\textwidth,  height=.30\textheight]{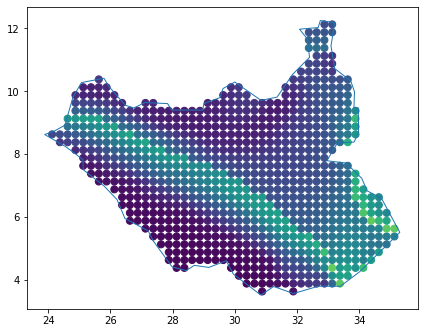}
    }
    \subfigure[Iran]{
        \includegraphics[width=0.45\textwidth, height=.30\textheight]{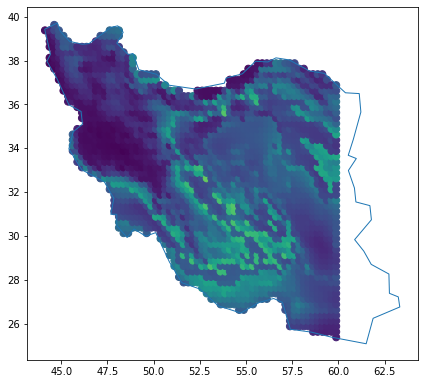}
    }
    \caption{Country-specific coefficient of variation (CV) maps showing fluctuations in the dominant aridity classification probabilities for Ethiopia, Morocco, South Sudan, and Iran during 1971–1989.}
    \label{fig:Country-specific Coefficient of Variability Map}
\end{figure}

To further explore sub-national patterns of fluctuation, we computed the coefficient of variation (CV) for individual pixels within four representative countries located in regions exhibiting high variability, namely Ethiopia (Horn of Africa), Morocco (Northwest Africa), South Sudan (East-Central Africa), and Iran (Middle East). These countries were selected due to their inclusion in prominent transition zones where arid, semi-arid, and non-arid classifications frequently shift. Figure \ref{fig:Country-specific Coefficient of Variability Map} illustrates the spatial distribution of CV values within these countries, highlighting localized hotspots of climate classification variability over the 1971–1989 period.

%%%%%%%%%%%%%%%%% IS %%%%%%%%%%%%%

\section{Discussion} \label{dis}
\subsection{Probability Interpretation}
Using a neural network classifier offers several advantages over a discrete categorization of climate zones. First, the model provides continuous probability estimates for each pixel, reflecting the likelihood of being classified as arid, semi-arid, or non-arid. This enables a more nuanced understanding of regional transitions and uncertainty in classification. For example, pixels with predicted probabilities like (Arid: $95\%$, Semi-Arid: $3.5\%$, Non-Arid: $1.5\%$) are most likely located in hyper-arid zones, such as the central Sahara or parts of the Rub' al Khali desert in Saudi Arabia. The extremely skewed distribution indicates high certainty in classification and high spatial homogeneity in these regions (Figure \ref{fig:Probability Map}).

On the other hand, a pixel with probabilities (Arid: $80\%$, Semi-Arid: $15\%$, Non-Arid: $5\%$) lies closer to ecological transition zones, possibly along the desert margins of the Sahel, southern Algeria, or northern Saudi Arabia. These pixels are still classified as desert but show increased uncertainty, signaling environmental gradients or changing land surface conditions. Finally, in transitional zones where probabilities are closer such as (Desert: $45\%$, Semi-Arid: $50\%$, Non-Arid: $5\%$), the uncertainty is substantial, suggesting the potential for classification shift over time or a climatologically mixed zone. The converse is also true. A pixel with a distribution like (Desert: $1\%$, Semi-Arid: $5\%$, Non-Arid: $94\%$) is highly stable and unlikely to transition categories. This behavior is captured well in the CV maps, which highlight temporal stability or fluctuation in classifications over time.

From the predicted probabilities, we derive discrete classifications by assigning each pixel to the category with the highest probability. This classification not only allows spatial mapping of climate zones but also helps identify climatic ecotones, or boundaries where sharp transitions occur. Regions with a high mix of probabilities tend to form gradual transitional zones, rather than sharp edges, which traditional discrete classifiers would miss. This illustrates the spatial continuity of ecological and climatic processes. For example, boundaries across the Sahel belt or the Iranian Plateau are revealed more accurately with probabilistic classifications, which respect the spatial heterogeneity in climate.

\subsection{Fluctuation Analysis: General Observations and Country-Specific Trends}

The coefficient of variation map (Figure \ref{fig:Coefficient of Variability}) highlight key regions of high inter-annual variability in classification probabilities. These include: (1) northwestern Africa, particularly Morocco and parts of Algeria; (2) the Horn of Africa, including Ethiopia and South Sudan; and (3) western Asia, particularly Iran. These areas show substantial temporal instability, with classifications frequently shifting among arid, semi-arid, and non-arid categories from year to year. These fluctuations can arise from seasonal anomalies, long-term climate variability, or changes in precipitation and temperature trends. Notably, variability is most pronounced in regions with diverse topography or microclimates \citep{ogunrinde2024long}. As such, high CV values point to ecological sensitivity, making these regions key targets for climate adaptation and land management planning.

To investigate these fluctuations further, we analyze average probabilities for the three categories across four representative countries: Morocco, Ethiopia, South Sudan, and Iran.

\begin{itemize}
\item Non-Arid Probabilities: South Sudan and Morocco exhibited the largest year-to-year variability. South Sudan’s non-arid probabilities ranged from over $70\%$ to below $30\%$, while Morocco’s fluctuated between $40\%$ and less than $10\%$. Iran’s non-arid values also varied widely ($60\%$ to under $30\%$), whereas Ethiopia showed slightly more stability with values between $30\%$ and $60\%$.

\item Semi-Arid Probabilities: These were relatively more stable across countries. South Sudan averaged $35.6\%$, Iran $28.4\%$, Ethiopia $22.7\%$, and Morocco $49.7\%$, although Morocco again displayed higher inter-annual fluctuation.

\item Desert Probabilities: These were somewhat stable but varied more than semi-arid estimates. Morocco had the highest average ($54.9\%$), followed by Ethiopia ($33.5\%$), Iran ($28.6\%$), and South Sudan ($17.3\%$).
\end{itemize}
These trends suggest that no single category dominates any of these countries consistently over time. Instead, there is an internal climatic balance, likely driven by complex interactions of elevation, precipitation, and regional weather systems.

Proportional averages further illustrate this:
\begin{itemize}
    
\item Ethiopia: 2:1 (non/semi-arid), 4:3 (non/arid), 2:3 (semi-arid/arid)

\item Morocco: 1:1 (non/semi-arid), 1:2 (non/arid), 1:2 (semi-arid/arid)

\item South Sudan: 5:3 (non/semi-arid), 5:2 (non/arid), 1:2 (semi-arid/arid)

\item Iran: 4:3 (non/semi-arid), 4:3 (non/arid), 1:1 (semi-arid/arid)

\end{itemize}
These ratios point to coexisting climate zones within each country, which explains why they are hotspots of climatic fluctuation. Transitional climate zones like the Moroccan coast, Iranian plateau, Ethiopian highlands, and South Sudan’s savannas often exhibit climatic variability \citep{lelieveld2016strongly}. For example, the subtropical dry zones in northwestern Africa transition quickly into semi-arid and arid zones inland. Similarly, Iran’s complex topography with mountain ranges flanking its central desert basin, creates pockets of variability across short spatial scales \citep{alijani2008developing}. In Ethiopia, elevation ranges from below sea level in the Danakil Depression to over 4,000 meters in the Simien Mountains. This altitudinal diversity produces a wide range of six agro-ecological zones in total, ranging from tropical to alpine climates \citep{MaharanaPyarimohan2018Ocvo}. South Sudan also experiences a topographic divide, with highlands surrounding central savannas and wetlands, creating stark differences in rainfall and vegetation cover \citep{SalihAbubakrA.M.2013Ciod}.

Finally, our results align with historical climatological studies. In Iran, studies from 1955 to 2000 found increasing maximum temperatures after the 1980s, while precipitation showed no significant trend \citep{KousariMohammadReza2011Mmam}. This temperature rise likely contributes to shifts in aridity classification, especially in semi-arid regions. \cite{ModarresReza2007Rtia} similarly found sporadic but significant changes in Iran’s arid and semi-arid zones, supporting our observed fluctuations. In Ethiopia, a U.S. Geological Survey analysis revealed declining precipitation in the southern and southeastern regions from the 1970s to early 2000s \citep{funk2012climate}. Our results show increased variability near these agricultural zones, particularly along the margins of the cropping regions. Historical droughts in the Sahel and Sahara during the 1970s to the 1980s brought global attention to the region’s climatic instability. Recent studies suggest these regions continue to face increasing desertification risks due to poverty, agriculture-dependent livelihoods, and warming trends \citep{HuangJianping2020Gdvt, LebelThierry2009Rtit, GangneronFabrice2022Paso}.

\section{Conclusion} \label{con}

In this study, we develop and implement a spatio-temporal feedforward neural network to classify arid, semi-arid, and non-arid zones across the Sahara and Sahel regions in Africa. The model demonstrates high accuracy in identifying arid and non-arid regions, when we assume the K\"oppen-Trewartha classification to be the ground truth. Additionally, our fluctuation analysis reveals transitional zones with high temporal variability, indicating the model’s sensitivity to shifts in climate categories and providing a novel way to quantify classification uncertainty at a pixel level.

Our approach of probabilistic classification provides insight into regions with strong versus weak probabilities for a particular climate categorization. Our study does not claim to detect ecological change or desertification; rather, our aim is to present a tool that can (1) interpolate arid and semi-arid classifications more flexibly and (2) provide an explicit probability/uncertainty estimate behind an established classification. Through this approach, we suggest that the boundaries between climate categories are more nuanced and dynamic than often assumed, and that a probabilistic perspective can reveal more complex relationships between climatic regions that may otherwise be overlooked when using simple deterministic categorizations.

Nonetheless, the study has several limitations. First, the model relies solely on historical climate variables and excludes vegetation indices or land-use data, which may enhance classification accuracy, particularly in heterogeneous semi-arid regions. Second, the use of a fixed feedforward architecture may constrain the model’s adaptability across different spatial or temporal contexts, and future work should explore more flexible architectures such as convolution or recurrent neural networks. Third, our model does not incorporate future climate projections, which limits its utility for long-term ecological forecasting and planning.

Finally, this approach relies on the assumption that incorporating spatio-temporal basis functions allows the model to sufficiently capture the underlying dependence structure in the data, so that the observations are treated as conditionally independent while evaluating the loss function. While this assumption simplifies model training, it may limit the model’s ability to account for complex spatial and temporal autocorrelation patterns that influence climate variability. As future work, a spatio-temporal feedforward classification neural network can be developed in which the spatio-temporal correlation structure is explicitly learned and optimized within the architecture itself. In addition, the use of static knot placements for the basis functions might miss localized patterns in highly dynamic areas. Adaptive knot placement or spatially-varying basis functions could potentially improve performance in transitional zones like the Sahel. Although we acknowledge limitations in our proposed methodology, we argue that introducing tools like feedforward neural networks into climate classification provides a promising avenue for future research, particularly in incorporating uncertainty into traditional deterministic systems.

\appendix

\section{Accuracy Metrics Over Time for Arid, Semi-Arid, and Non-Arid Zones} \label{app1}

\begin{figure}[H]
    \centering
    \includegraphics[width=0.75\linewidth]{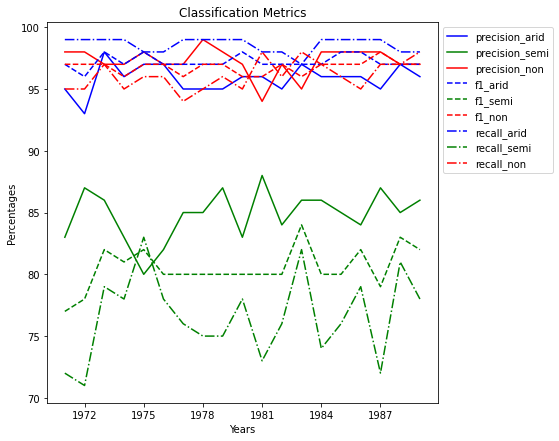}
    \caption{Classification performance metrics (precision, recall, and F1-score) over time for arid (blue), semi-arid (green), and non-arid (red) climate zones. Metrics are expressed as percentages.}
    \label{metricstime}
\end{figure}

Figure \ref{metricstime} shows the classification metrics for the test set years, that is, from the early 1970s through the late 1980s. The precision and recall for the arid and non-arid zones are consistently high, mostly above 95\% with minor fluctuations across years. In contrast, the precision and recall metrics for the semi-arid zone exhibit greater variability, with values generally ranging between 70\% and 85\%. This elevated variability is expected, as the semi-arid class covers substantially fewer pixels than the arid and non-arid zones, making its precision and recall more sensitive to small classification errors.

The F1-scores follow similar patterns; they remain stable and high for arid and non-arid zones, indicating robust performance, while the semi-arid F1-score shows larger variations. Overall, these trends suggest that while the model consistently achieves high accuracy in the dominant arid and non-arid regions, classification of the smaller semi-arid areas is more challenging and uncertain over time because of significant climatic flux in these transitional zones, which leads to greater variability and uncertainty in classification.

\subsection*{Disclosure statement} 
The authors report there are no competing interests to declare.

%\subsection*{Data availability statement}
%All datasets used in this study are publicly available through the NASA Global Land Data Assimilation System Version 2 (GLDAS-2). Specifically, the data can be accessed at:
%\url{https://disc.gsfc.nasa.gov/datasets/GLDAS_NOAH025_M_2.0/summary} (Accessed: July 08, 2025).

\bibliographystyle{plainnat}
\bibliography{Reference}
\end{document}